\documentclass{article}
\PassOptionsToPackage{nonamebreak}{natbib}
\usepackage{iclr2027_conference,times}

\usepackage{amsmath,amsfonts,bm}

\def\eqref#1{equation~\ref{#1}}

\def\1{\bm{1}}

\DeclareMathAlphabet{\mathsfit}{\encodingdefault}{\sfdefault}{m}{sl}
\SetMathAlphabet{\mathsfit}{bold}{\encodingdefault}{\sfdefault}{bx}{n}

\usepackage{xcolor}
\definecolor{ars}{HTML}{B20000}
\definecolor{robometer}{HTML}{76A3A0}
\definecolor{soler1}{HTML}{95A06A}
\definecolor{lrm}{HTML}{6FD999}
\definecolor{r2vlm}{HTML}{9C92C0}
\definecolor{topreward}{HTML}{BE8D96}

\newcommand{\arscolor}[1]{\textcolor{ars}{#1}}
\newcommand{\robometercolor}[1]{\textcolor{robometer}{#1}}
\newcommand{\solercolor}[1]{\textcolor{soler1}{#1}}
\newcommand{\lrmcolor}[1]{\textcolor{lrm}{#1}}
\newcommand{\rtwovlmcolor}[1]{\textcolor{r2vlm}{#1}}
\newcommand{\toprewardcolor}[1]{\textcolor{topreward}{#1}}

\usepackage{float}
\usepackage{hyperref}
\usepackage{url}
\usepackage{graphicx}
\usepackage{capt-of}
\usepackage{booktabs}
\usepackage{multirow}
\usepackage{amsmath}
\usepackage{amssymb}
\usepackage{bbm}
\usepackage{enumitem}

\title{ARS: Agentic Reward System for Robot Learning}

\author{%
Sheng Hu
\And
Weiyi Lu
\And
Lingbing Zeng
\And
Gan Weng
\AND
Weiwei Zhang
\And
Kai Xie
\And
Xiaofeng Mou
\And
Yi Xu\thanks{Corresponding author.}
}

\iclrfinalcopy
\begin{document}

\maketitle
\lhead{Preprint}
\renewcommand{\headrulewidth}{0.4pt}

\vspace{-1.5em}
{\centering\normalfont
AIRC, Midea Group\\
{\fontsize{8}{10}\selectfont\texttt{\{husheng24,luwy54,zenglb8,wenggan,zhangww44,xiekai17,mouxf,xuyi42\}@midea.com}}\par}
\vspace{1.5em}

\begin{abstract}
Progress reward modeling is the problem of estimating how a robot's behavior changes task progress over time. Reliable estimation requires distinguishing meaningful state changes from failed attempts and task-irrelevant actions. We introduce the \textbf{Agentic Reward System (ARS)}, an inference framework for progress reward modeling with general-purpose vision-language models (VLMs), without additional reward-model training. Given an offline trajectory and a task instruction, ARS uses adaptive visual inspection for both event proposal and verification. A subagent proposes a task-relevant event timeline, which a primary agent verifies and revises before estimating per-frame progress. ARS can incorporate optional terminal outcome labels and visual references to inform its judgments. It can also audit progress estimates from external reward models. We evaluate ARS with a 27B VLM on a controlled semantic-mismatch benchmark and downstream policy learning in simulation and on a real robot. The benchmark reveals that several evaluated reward baselines assign spurious progress to wrong-object manipulation even in simple pick-and-place scenes. ARS better suppresses these errors and outperforms these baselines in simulation policy learning. We further demonstrate that ARS supports long-horizon policy learning from mixed-quality offline experience on real-robot multi-screw fastening in a full-scale laboratory replica of an industrial washing-machine assembly line. These results suggest that structured inference and verification can improve the usefulness of general-purpose VLMs for robot reward modeling. Code is at \url{https://github.com/midea-ai/ars}.
\end{abstract}

\begin{figure}[t]
    \centering
    \setlength{\abovecaptionskip}{6pt}
    \includegraphics[width=\linewidth]{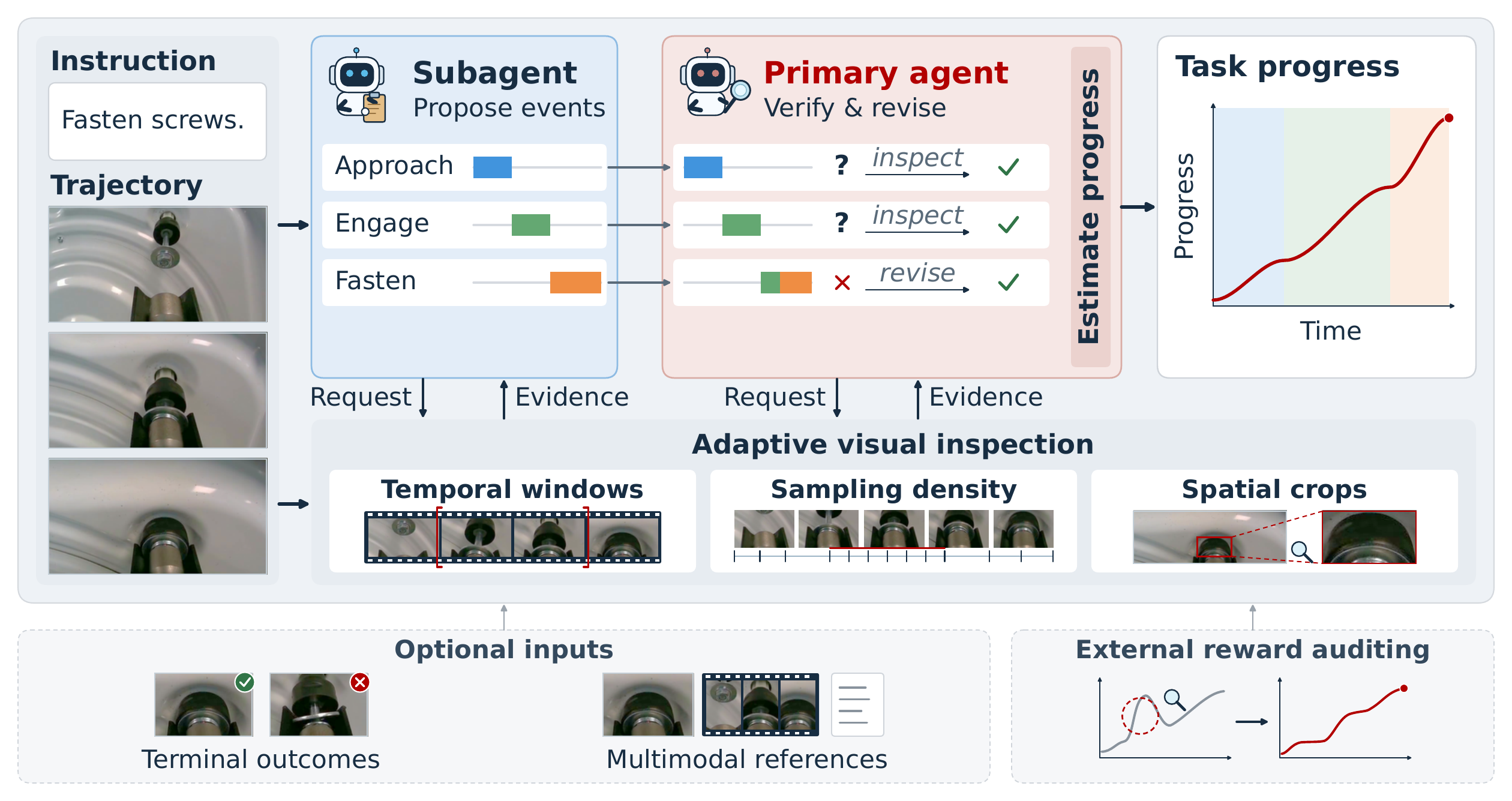}
    \caption{Overview of ARS. ARS uses adaptive visual inspection for both event proposal and verification. A subagent proposes task-relevant events from an instruction and offline trajectory. The primary agent verifies and refines these proposals, then estimates per-frame task progress. ARS supports optional terminal outcomes and multimodal references, plus external reward auditing.}
    \label{fig:ars-overview}
\end{figure}

\section{Introduction}

Reliable rewards are central to learning robot policies from experience. Sparse terminal feedback offers limited guidance about individual actions, while hand-designed dense rewards require task-specific engineering and often privileged simulator state. In long-horizon manipulation, successful trajectories can contain failed attempts, while failed trajectories can retain useful intermediate actions. Estimating task progress from visual observations can identify these local differences and turn mixed-quality experience into training supervision.

Vision-language models (VLMs) can interpret task instructions together with visual trajectories. Existing approaches elicit progress from pretrained models \citep{gvl,topreward} or train dedicated reward models using trajectory supervision \citep{robometer,lrm,robo-dopamine}. Prior studies show that specialized robotic reward models can outperform directly prompted frontier flagship VLMs of the time, including GPT and Gemini \citep{r2vlm,soler1,roboreward}. These findings raise the question of how much a general-purpose VLM's reward capabilities depend on the inference procedure used to obtain its judgments.

Accurate estimation requires checking the evidence behind an interpretation: smooth manipulation of the wrong object does not advance the instructed task, and an apparent grasp may require later observations for confirmation. Relevant evidence varies in temporal extent and spatial location. Recent methods structure estimation through spatiotemporal, recursive, and recurrent reasoning \citep{soler1,rover-recursive,r2vlm}. We build on this direction by letting unresolved event claims guide additional visual inspection and revisions to an explicit event timeline.

We introduce the \textbf{Agentic Reward System (ARS)} (Figure~\ref{fig:ars-overview}). A subagent proposes task-relevant events and their intervals; a primary agent verifies and revises them. Both agents adaptively select temporal windows, sampling densities, and spatial crops to inspect visual evidence. The primary agent then constructs per-frame progress from verified state changes, assigning no additional credit to failed attempts and allowing regressions after setbacks (Figure~\ref{fig:ars-timeline}). ARS requires no additional reward-model training and operates on offline trajectories, allowing later evidence to resolve earlier ambiguities. The workflow can optionally use terminal outcomes and visual references, or audit external reward predictions.

We evaluate ARS using a 27B general-purpose VLM. We construct a controlled semantic-mismatch benchmark in ManiSkill \citep{maniskill}, revealing that several evaluated reward systems assign spurious progress under wrong-object instructions even in simple pick-and-place scenes with only two candidate objects. ARS better suppresses these errors. On the long-horizon manipulation tasks of LIBERO-Long \citep{libero}, selecting action chunks by estimated progress improves policy success by 13.7 percentage points over unfiltered behavior cloning (BC) and 10.3 points over the strongest compared reward baseline. On real-robot multi-screw fastening in a full-scale laboratory replica of an industrial washing-machine assembly line, ARS selection with visual references raises success from 58.0\% for demonstration-only behavior cloning to 81.5\%. Together, these results highlight the importance of the inference workflow used to elicit and verify reward judgments.

Our contributions are:
\begin{itemize}[leftmargin=*,nosep]
    \item We introduce ARS, combining adaptive visual inspection, an editable event timeline, and cross-agent verification for progress estimation without reward-model training.
    \item We construct a controlled semantic-mismatch benchmark that exposes spurious credit for wrong-object manipulation in simple pick-and-place scenes.
    \item We show that ARS with general-purpose VLMs outperforms the evaluated specialized reward models and supports long-horizon policy learning in simulation and on an industrially relevant real-robot task.
\end{itemize}

\section{Related Work}

\paragraph{Robotic reward and progress estimation.}
Robotic reward estimation draws on pretrained visual representations \citep{vip,liv}, video--language similarity \citep{roboclip}, temporal supervision \citep{rank2reward,rewind,timerewarder}, and VLM-generated preferences \citep{rl-vlm-f}. Dedicated models learn task-conditioned progress and completion signals \citep{roboreward,robo-dopamine,robometer,lrm,vlac,densereward}, while GVL and TOPReward elicit progress without reward-model training through shuffled-frame inference and token probabilities, respectively \citep{gvl,topreward}. Other approaches exploit stage structure \citep{sarm,victor,stdr} and relative progress \citep{arm,warp-rm}, or improve reliability through label correction, online refinement, confidence gating, and prompt optimization \citep{ur-vc,progressor,rarm,demo2reward}. Explicit reasoning further structures estimation. Schroeder et al. introduced ROVER \citep{rover-recursive} to structure training-free progress estimation through recursive subtask reasoning. They later proposed SOLE-R1 \citep{soler1}, a video-language reward model trained for spatiotemporal reasoning and dense progress prediction. ProcVLM uses procedural reasoning \citep{procvlm}, while ProgressLM combines demonstration-conditioned episodic retrieval with mental simulation \citep{progresslm}. R\textsuperscript{2}VLM recurrently updates task decomposition and completion status \citep{r2vlm}. ARS builds on these directions by letting unresolved event claims guide additional visual inspection and revisions to an event timeline.

\paragraph{Agentic inference and reward verification.}
Reward models support action selection, execution auditing, and world-model planning \citep{rover,tapsampling,prm-as-a-judge,ev-wm}. Agentic inference provides mechanisms for acquiring and checking the evidence underlying such judgments. ReAct interleaves reasoning and tool use \citep{react}, while VideoAgent adaptively retrieves visual information \citep{videoagent}. Agentic reward methods apply related ideas to evaluation: ARM-Thinker invokes multimodal tools \citep{arm-thinker}; ProRe coordinates a reasoner with evaluator agents \citep{prore}, while the Agentic Verifier framework introduced with AgentV-RL combines forward and backward verification agents \citep{agentv-rl}; VAGEN inspects trajectories and probes GUI environments \citep{vagen}. Reward as an Agent uses structured evaluation and reflection for generated embodied videos \citep{reward-as-agent}. ARS applies active verification to executed robot trajectories, coupling event proposals with verification in separate reasoning contexts. Its harness requires evidence coverage and renewed confirmation after event revisions before dense progress is constructed.

\section{Agentic Reward System}

\subsection{Problem setup}

Let $\mathcal D$ be an offline dataset of robot trajectories. For a trajectory $\tau\in\mathcal D$, we write $\tau=(\mathbf{o}_{0:T},\mathbf{a}_{0:T-1},l)$. Here, $\mathbf{o}_t$ and $\mathbf{a}_t$ denote the observation and action at time $t$, while $l$ is the task instruction and $T$ is the terminal index.

A reward system $f$ typically takes an observation sequence and a task instruction as input and predicts per-frame task progress:

\begin{equation}
    p_{0:T}
    = f\!\left(\mathbf{o}_{0:T},l\right)
    \in[0,1]^{T+1}.
    \label{eq:problem-output}
\end{equation}

Estimating progress at time $t$ may require observations beyond $\mathbf{o}_{0:t}$ to resolve ambiguity about the task state. For example, whether an apparent grasp has secured an object may only become clear when the object is subsequently lifted or transported. The temporal location and extent of the evidence needed for verification vary across events, motivating adaptive visual inspection guided by unresolved questions about the trajectory.

\begin{figure}[!t]
    \centering
    \includegraphics[width=\linewidth]{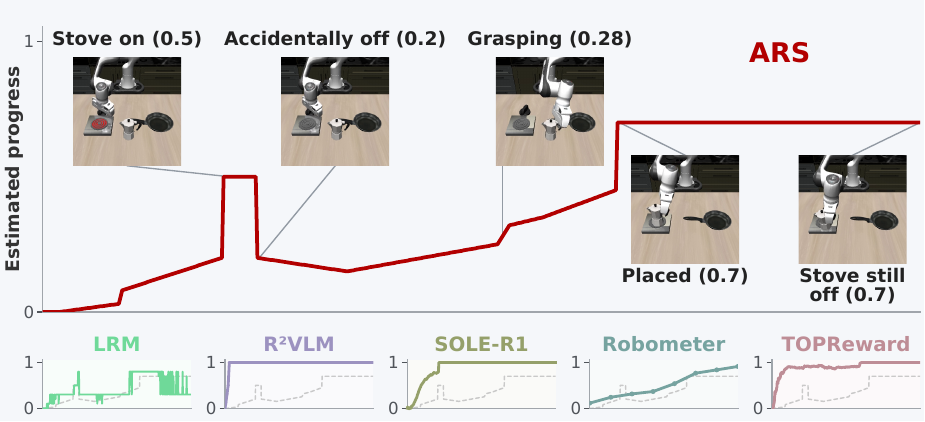}
    \caption{Progress estimates for a partially successful LIBERO trajectory under the instruction ``turn on the stove and put the moka pot on it.'' Insets show selected frames and \arscolor{ARS} progress values. ARS captures the regression when the stove is unintentionally switched off and credits subsequent progress toward pot placement. Baseline curves appear below (Appendix~\ref{app:compared-methods}), with ARS repeated as a light gray dashed line for reference. The event proposal and primary-agent verification for this trajectory are detailed in Appendix~\ref{app:ars-reasoning-trace}.}
    \label{fig:ars-timeline}
\end{figure}

\subsection{Adaptive visual inspection}

We observe that agents given a complete trajectory video often produce coarse or incorrect interpretations, particularly for long-horizon tasks. Brief state transitions can be overlooked, and unsuccessful attempts can be conflated with later successful interactions. Access to all frames therefore does not ensure close examination of the evidence relevant to task progress.

ARS supports adaptive visual inspection through tool calls. An agent selects a temporal window, adjusts its sampling density, and optionally requests a spatial crop. The request is guided by the current question and can extend to later observations when an earlier event requires retrospective verification. The returned visual evidence informs subsequent reasoning and inspection requests. Our workflow ablations show that targeted inspection provides limited additional gains on the short semantic-mismatch benchmark, while improving downstream policy performance on long-horizon tasks even when the complete video is provided initially (Section~\ref{sec:ablations}).

\subsection{Event-grounded reasoning}

ARS represents the trajectory as a timeline of task-relevant events. Each event associates a temporal interval with a description grounded in cited visual evidence. The harness maintains the event timeline and its supporting evidence across reasoning contexts. Agent-authored state updates are executed through tool calls. When an event is created or revised, its cited evidence must cover the full event interval. Before proposal generation can finish, the event collection must cover the complete trajectory. Failed checks return the missing intervals for correction, making incomplete coverage an actionable part of inference.

\subsection{Cross-agent verification}
\label{sec:cross-agent-verification}

A single agent can complete the workflow by inspecting and revising events before estimating progress. However, ARS delegates event proposal to a subagent, which constructs a task-relevant event timeline through adaptive visual inspection. In a separate reasoning context, the primary agent gathers additional visual evidence to verify and revise these proposals.

Event confirmation requires evidence acquired during primary-agent verification. The harness checks that sampled frames include the event interior and the preceding and following context available within the trajectory. Revising an event invalidates its confirmation, requiring the updated claim to be checked again. These checks enforce evidence requirements; the primary agent judges whether the visual content supports the event description and its timing.

The primary agent then estimates per-frame progress from the verified events and their supporting evidence. The two agents may share the same underlying VLM.

This organization gives the primary agent explicit claims to challenge. For long-horizon tasks, this separation also reduces the primary agent's context usage by retaining event proposals without carrying over the subagent's full interaction history, leaving more capacity for targeted verification and progress estimation. See Section~\ref{sec:ablations} for further analysis.

\subsection{Progress estimation}

A central goal of progress-based reward modeling is to translate task understanding into useful numerical estimates of progress. Training-free methods approach this through numerical prediction over shuffled frames (GVL), completion-token probabilities (TOPReward), or recursive subtask reasoning (ROVER) \citep{gvl,topreward,rover-recursive}. ARS grounds progress estimation in an explicit timeline of task events whose descriptions and timing undergo targeted visual inspection and independent verification. The primary agent constructs and revises the progress curve using these verified events and their supporting evidence. Figure~\ref{fig:ars-timeline} illustrates how the resulting estimates capture local regression while preserving partial task achievement.

Through tool calls, the agent chooses between interpolated segments and local pointwise edits. Progress readback and summary tools expose the resulting curve's shape and explicitly flag jump locations and magnitudes, allowing the agent to check its estimates against visual evidence and revise them. Appendix~\ref{app:agent-prompts-tools} summarizes the agent prompts and available tools.

\subsection{Flexible Integration of Auxiliary Information}
\label{sec:aux}

As an agentic system, ARS can incorporate auxiliary context and audit external reward proposals within the same reasoning and verification workflow. Terminal outcomes and references help clarify task criteria that may be ambiguous from visual observations and brief instructions.

\begin{itemize}[leftmargin=*,nosep]
    \item \textbf{Terminal outcomes.} Trajectory-level success/failure labels provide low-cost supervision that may already be available from data collection. ARS uses them to anchor terminal completion while estimating intermediate progress from verified events, preserving local regressions in successful trajectories and partial progress in failed ones. We evaluate outcome conditioning in downstream policy learning (Section~\ref{sec:bc}, Table~\ref{tab:libero}).
    \item \textbf{In-context references.} Building on in-context progress estimation \citep{gvl,vlac}, ARS accepts images, video clips, textual descriptions, and positive or negative examples. Progress annotations are optional, allowing references to clarify task criteria without numerical labels. Section~\ref{sec:realrobot} examines their use in real-world policy learning.
    \item \textbf{External reward auditing.} ARS audits progress curves from external reward models, using questionable predictions to guide visual inspection and revising the curves according to verified task events. Section~\ref{sec:semantic-mismatch} evaluates this capability. The corrected curves and accompanying audit traces could also be explored as potential supervision for reward-model training.
\end{itemize}

\subsection{Progress-guided policy learning}
\label{sec:progress-guided-policy-learning}

ARS is compatible with value learning from progress-derived rewards \citep{robometer,robo-dopamine}. However, to evaluate the contribution of its progress estimates more directly, we adopt progress-guided policy learning following prior work on weighted behavior cloning \citep{vip,gvl,sarm,warp-rm,arm}. We use local progress changes to determine whether an action chunk contributes to policy training. This makes the mapping from progress estimates to training supervision explicit and avoids introducing an additional critic's estimation errors and optimization choices into the comparison.

We measure progress over chunks of $H$ consecutive actions. For a chunk starting at $t$, we consider the trajectory segment $(\mathbf{o}_{t:t+H},\mathbf{a}_{t:t+H-1},l)$ and define its progress-derived score as
\begin{equation}
    \Delta_t
    = p_{t+H} - p_t.
    \label{eq:progress-chunk-score}
\end{equation}

We introduce an indicator function $I_t$ for chunk selection, defined as
\begin{equation}
    I_t
    = \mathbbm{1}\!\left[\Delta_t>\epsilon(p_t)\right]\in\{0,1\},
    \label{eq:progress-chunk-weight}
\end{equation}
where $\epsilon(p_t)$ is a threshold that depends on the current progress value $p_t$.

Over the dataset $\mathcal D$, we minimize the following binary-weighted behavior-cloning objective:
\begin{equation}
    \mathcal L_{\mathrm{BC}}(\theta)
    =
    \frac{1}{\sum_{\tau\in\mathcal D}\sum_t I_t}
    \sum_{\tau\in\mathcal D}\sum_t I_t\,
    \ell_{\mathrm{BC}}\!\left(
        \theta;
        \mathbf o_t,l,\mathbf a_{t:t+H-1}
    \right).
    \label{eq:bc}
\end{equation}
Here, $\theta$ denotes the policy parameters, and $\mathbf a_{t:t+H-1}$ is the recorded action sequence. The function $\ell_{\mathrm{BC}}$ denotes the policy's per-chunk behavior-cloning loss, conditioned on the current observation $\mathbf o_t$ and instruction $l$. The sums range over trajectories $\tau\in\mathcal D$ and their candidate chunk starting positions $t$, with the weights and action sequences defined for the corresponding trajectory.

We choose one of two alternative rules for the threshold function $\epsilon(p_t)$:

\begin{itemize}[leftmargin=*,nosep]
    \item \textbf{Zero threshold.} In the simplest setting, we set $\epsilon(p_t)=0$ for all chunks, so that only chunks with positive estimated progress contribute to policy training.
    \item \textbf{Stage-dependent threshold.} Inspired by the advantage thresholding in $\pi_{0.6}^{*}$ \citep{pi06}, we choose thresholds to retain the top fraction $\alpha$ of chunks ranked by $\Delta_t$. Global ranking can favor early stages or stages with larger progress changes. We divide the progress range $[0,1]$ into $B$ equal-width bins and assign each chunk to a bin according to its current progress value $p_t$. Within each bin, we rank chunks by their progress changes $\Delta_t$ and set $\epsilon(p_t)$ to a shared threshold that retains the top fraction $\alpha$.
\end{itemize}

\section{Experiments}

We evaluate task alignment through controlled semantic mismatches and progress-guided policy learning on long-horizon simulation and real-robot tasks. Ablations examine base-model choice and inference components.

\paragraph{Compared methods and experimental settings.}
We compare \arscolor{ARS} against \lrmcolor{LRM} \citep{lrm}, \rtwovlmcolor{R\textsuperscript{2}VLM} \citep{r2vlm}, \solercolor{SOLE-R1} \citep{soler1}, \robometercolor{Robometer} \citep{robometer}, and \toprewardcolor{TOPReward} \citep{topreward}. Implementation details of these methods are provided in Appendix~\ref{app:compared-methods}.
Although ARS can run with proprietary models, we use open-weight Qwen3.6-27B \citep{qwen36} in our main experiments for reproducibility.
The main reward-system comparisons use only the trajectory and instruction as inputs. Configurations using auxiliary information or privileged simulator state are identified separately.

\subsection{Semantic-mismatch benchmark}
\label{sec:semantic-mismatch}

Related task-alignment evaluations use cross-task video--instruction pairing \citep{roboclip,rewind,robometer,topreward}, counterfactual relabeling \citep{roboreward}, and distractor instructions \citep{r2vlm}. We isolate wrong-object credit assignment in simple pick-and-place tasks, measuring the temporal extent of spurious progress increases and peak false progress.

\paragraph{Task and data.}
Inspired by prior work \citep{rl4vla}, we generate motion-planner expert trajectories in ManiSkill \citep{maniskill}, with two candidate objects per tabletop scene and randomized backgrounds. Each video is paired with the correct pick-and-place instruction and a mismatch referring to the other visible object. Across 25 object types, the benchmark contains 500 pairs (1,000 samples), averaging 31.7 frames per trajectory. Expert trajectories advance the correct task at every recorded step; privileged simulator states verify that the gripper never approaches the mismatch object, even when camera projection suggests proximity. Training-data exposure is summarized in Appendix~\ref{app:maniskill-exposure}.

\paragraph{Metrics.}
We evaluate using three complementary metrics:
\begin{itemize}[leftmargin=*,nosep]
    \item \textbf{Value-Order Correlation (VOC)} \citep{gvl} reports the mean per-trajectory Spearman rank correlation between predicted progress and timestep under correct instructions.
    \item \textbf{False Progress Suppression Ratio (FPSR)} measures the fraction of mismatch-trajectory duration not covered by strictly positive progress-increase intervals. See Appendix~\ref{app:fpsr} for details.
    \item \textbf{Correct--Mismatch Progress Gap (CMPG)} is the difference between the mean terminal progress under correct instructions and the mean per-trajectory maximum progress under mismatch instructions. See Appendix~\ref{app:cmpg} for details.
\end{itemize}

\begin{table}[t]
\caption{Results on the semantic-mismatch benchmark. The Audit configuration uses Robometer predictions as an external proposal, audited and revised by ARS with Qwen3.6-27B.}
\label{tab:mismatch-main}
\begin{center}
\normalsize
\setlength{\tabcolsep}{4pt}
\begin{tabular*}{\linewidth}{@{\extracolsep{\fill}}llccc@{}}
\toprule
Group & Method & VOC $\uparrow$ & FPSR $\uparrow$ & CMPG $\uparrow$ \\
\midrule
\multirow{4}{*}{Reward-trained}
 & \lrmcolor{LRM} & 0.828 & 0.893 & 0.373 \\
 & \rtwovlmcolor{R\textsuperscript{2}VLM} & 0.916 & 0.184 & -0.050 \\
 & \solercolor{SOLE-R1} & 0.902 & 0.361 & 0.266 \\
 & \robometercolor{Robometer} & 0.983 & 0.240 & 0.366 \\
\midrule
Audit & \arscolor{ARS} w/ \robometercolor{Robometer} proposal & 0.988 & 0.915 & 0.891 \\
\midrule
\multirow{3}{*}{Reward-training-free} & \toprewardcolor{TOPReward} + Qwen3-VL-8B & 0.930 & 0.368 & 0.685 \\
 & \toprewardcolor{TOPReward} + Qwen3-VL-32B & 0.945 & 0.383 & 0.756 \\
 & \arscolor{ARS} + Qwen3.6-27B & \textbf{0.995} & \textbf{0.918} & \textbf{0.904} \\
\bottomrule
\end{tabular*}
\end{center}
\end{table}

Table~\ref{tab:mismatch-main} shows that temporal consistency does not ensure instruction alignment: Robometer achieves 0.983 VOC but only 0.240 FPSR, while R\textsuperscript{2}VLM assigns higher mean peak progress to mismatched instructions than mean terminal progress to correct ones (negative CMPG). ARS leads all three metrics without reward-model training, combining accurate ordering with suppression of spurious progress. Additional TOPReward base-model results appear in Appendix~\ref{app:topreward-base-model}.

Auditing Robometer predictions with ARS raises FPSR from 0.240 to 0.915 and CMPG from 0.366 to 0.891 while maintaining high VOC (0.988). These results approach ARS without proposals, supporting its use to correct external reward predictions.

\subsection{Policy learning in simulation}
\label{sec:bc}

\paragraph{Offline rollout dataset.}
A fixed behavior policy collects 4,096 trajectories across the complete LIBERO-Long suite \citep{libero}: 2,204 successful and 1,892 failed trajectories, averaging 390.3 frames. All reward systems score this shared dataset, whose successive object interactions provide mixed-quality chunks within both successful and failed trajectories. Training-data exposure and task instruction clarifications are detailed in Appendices~\ref{app:libero-exposure} and~\ref{app:task-instruction-clarifications}.

\paragraph{Training protocol.}
All downstream policies start from the same pretrained $\pi_{0.5}$ checkpoint \citep{pi05}, with chunk horizon $H=10$ and a shared training and evaluation schedule. The behavior policy serves only data collection. We apply the zero-threshold rule (Section~\ref{sec:progress-guided-policy-learning}) to each reward system's estimates, retaining chunks with positive progress differences at unit weight; Vanilla BC retains all chunks. The retained ratio is the percentage of candidate chunks kept. Appendix~\ref{app:bc-training-evaluation} details the protocol and selection rationale.

\begin{table}[t]
\caption{Policy-learning results on LIBERO-Long. Retained is the micro-averaged percentage of action chunks kept from the original dataset; gains are in percentage points.}
\label{tab:libero}
\begin{center}
\normalsize
\setlength{\tabcolsep}{3pt}
\begin{tabular*}{\linewidth}{@{\extracolsep{\fill}}llccc@{}}
\toprule
Group & Method & Success (\%) $\uparrow$ & vs.\ Vanilla $\uparrow$ & Retained (\%) \\
\midrule
Baseline & Vanilla BC & 44.2 $\pm$ 0.3 & {---} & 100.0 \\
\midrule
\multirow{4}{*}{Reward-trained}
 & \lrmcolor{LRM} & 38.6 $\pm$ 0.5 & -5.6 & 13.7 \\
 & \rtwovlmcolor{R\textsuperscript{2}VLM} & 41.5 $\pm$ 0.4 & -2.7 & 13.6 \\
 & \solercolor{SOLE-R1} & 44.8 $\pm$ 0.4 & +0.6 & 52.6 \\
 & \robometercolor{Robometer} & 46.8 $\pm$ 0.9 & +2.6 & 89.5 \\
\midrule
\multirow{3}{*}{Reward-training-free} & \toprewardcolor{TOPReward} + Qwen3-VL-8B & 43.1 $\pm$ 0.5 & -1.1 & 77.0 \\
 & \toprewardcolor{TOPReward} + Qwen3-VL-32B & 47.6 $\pm$ 0.7 & +3.4 & 67.9 \\
 & \arscolor{ARS} + Qwen3.6-27B & \textbf{57.9} $\pm$ 0.4 & \textbf{+13.7} & 66.9 \\
\specialrule{0.5pt}{0.8ex}{0pt}
\specialrule{0.5pt}{2pt}{0.8ex}
\multirow{2}{*}{Outcome labels} & Success-only BC & 52.6 $\pm$ 0.7 & +8.4 & 38.5 \\
 & \arscolor{ARS} + outcome & 62.2 $\pm$ 0.6 & +18.0 & 60.3 \\
\midrule
Privileged state & Privileged progress & 66.5 $\pm$ 0.8 & +22.3 & 57.2 \\
\bottomrule
\end{tabular*}
\end{center}
\end{table}

ARS reaches 57.9\% success, the highest among systems using only trajectories and instructions, exceeding Vanilla BC by 13.7 points (Table~\ref{tab:libero}). Without reward-model training, it also surpasses LRM, SOLE-R1, and Robometer, which have documented LIBERO training exposure. ARS and TOPReward-32B retain similar proportions of chunks (66.9\% and 67.9\%), yet differ by 10.3 points in success, supporting the importance of selection quality.

LRM and R\textsuperscript{2}VLM retain fewer than 14\% of chunks and underperform Vanilla BC. Their quantized or repeated progress predictions yield many zero increments, limiting temporal information for selection (Appendix~\ref{app:progress-diagnostics}).

\paragraph{Outcome conditioning.}
ARS + outcome additionally receives a terminal success/failure label for each trajectory during progress inference. It reaches 62.2\% success, 9.6 points above success-only BC (which retains all chunks from trajectories labeled successful) and 4.3 points below the privileged-state reference (66.5\%), capturing 80.7\% of the reference's success-rate gain over Vanilla BC (44.2\%). Appendix~\ref{app:privileged-progress} describes how we construct the privileged progress signal from simulator state. Even with outcome conditioning, trajectories judged successful can retain flat or decreasing progress segments, while those judged unsuccessful can retain increasing segments (Appendix~\ref{app:progress-diagnostics}).

\subsection{Ablation study}
\label{sec:ablations}

\paragraph{Base model and inference procedure.}
Under the same ARS workflow, Qwen3.6-27B substantially improves FPSR and CMPG over Qwen3.5-4B \citep{qwen35}, showing that progress estimation benefits from the capabilities of the base model (Table~\ref{tab:ablations-available}). Nevertheless, ARS with Qwen3.5-4B already exceeds Robometer-4B on all three metrics without additional reward-model training, despite using a different base model at the same nominal parameter scale. With Qwen3.6-27B held fixed, GVL-style single-prompt inference achieves substantially lower VOC and CMPG than ARS (Appendix~\ref{app:gvl-single-prompt}). Thus, the gains depend on how the model is used, beyond the choice of base model alone. Additional base-model execution details are provided in Appendices~\ref{app:base-model-execution}--\ref{app:small-model-long-horizon}.

\begin{table}[t]
\caption{Base-model and inference-procedure comparisons on the semantic-mismatch benchmark.}
\label{tab:ablations-available}
\begin{center}
\normalsize
\setlength{\tabcolsep}{4pt}
\begin{tabular*}{\linewidth}{@{\extracolsep{\fill}}lccc@{}}
\toprule
Configuration & VOC $\uparrow$ & FPSR $\uparrow$ & CMPG $\uparrow$ \\
\midrule
\robometercolor{Robometer}-4B & 0.983 & 0.240 & 0.366 \\
\arscolor{ARS} + Qwen3.5-4B & 0.986 & 0.625 & 0.633 \\
GVL-style + Qwen3.6-27B & 0.637 & 0.900 & 0.354 \\
\arscolor{ARS} + Qwen3.6-27B & \textbf{0.995} & \textbf{0.918} & \textbf{0.904} \\
\bottomrule
\end{tabular*}
\end{center}
\end{table}

\paragraph{Inference usage and workflow ablations.}
ARS achieves higher downstream success than frame-wise TOPReward with substantially fewer model queries and cumulative input tokens, while introducing generative reasoning overhead (Table~\ref{tab:ablations-libero}). Compared with TOPReward using Qwen3-VL-32B, full ARS improves success from 47.6\% to 57.9\%, with 93.6\% fewer model queries and 65.8\% fewer cumulative input tokens.

With Qwen3.6-27B fixed, visual inspection improves success from 45.4\% to 52.3\%. Adding same-context self-verification yields a slightly lower mean success rate of 51.8\%, despite substantially higher inference usage. Full ARS verifies proposals in a separate reasoning context and reaches 57.9\%, outperforming the same-context control by 6.1 percentage points. These results support the benefit of separating proposal generation from verification in this setting. Appendix~\ref{app:workflow-ablations} details the workflow variants, inference usage, and additional semantic-mismatch results.

We further examine the role of explicit language chain-of-thought (CoT) in intermediate reasoning by disabling the base model's thinking mode; the results and analysis are provided in Appendix~\ref{app:intermediate-reasoning}.

\begin{table}[t]
\caption{Workflow ablations on LIBERO-Long. Input/output token counts are in thousands (K).}
\label{tab:ablations-libero}
\begin{center}
\normalsize
\setlength{\tabcolsep}{2pt}
\begin{tabular*}{\linewidth}{@{\extracolsep{\fill}}p{0.38\linewidth}rrrrr@{}}
\toprule
Configuration & Success (\%) $\uparrow$ & Retained (\%) & Queries & Input (K) & Output (K) \\
\midrule
\toprewardcolor{TOPReward} + Qwen3-VL-32B & 47.6 $\pm$ 0.7 & 67.9 & 390.3 & 2,866.3 & 0.0 \\
\midrule
\arscolor{ARS} w/o verification \& inspection & 45.4 $\pm$ 0.3 & 78.1 & 7.3 & 242.7 & 10.1 \\
\arscolor{ARS} w/o verification & 52.3 $\pm$ 0.8 & 70.3 & 10.2 & 413.2 & 9.9 \\
\arscolor{ARS} w/o context separation & 51.8 $\pm$ 0.5 & 68.7 & 19.0 & 1,097.2 & 15.2 \\
Full \arscolor{ARS} & \textbf{57.9} $\pm$ 0.4 & 66.9 & 24.9 & 980.6 & 21.1 \\
\bottomrule
\end{tabular*}
\end{center}
\end{table}

\subsection{Real-world policy learning}
\label{sec:realrobot}

We evaluate sequential fastening of three screws on a washing machine's rear panel in a full-scale laboratory replica of an industrial assembly line (Figure~\ref{fig:realrobot-setup}; Appendix~\ref{app:realrobot-details}).

\paragraph{Experimental setup.}
We use 2,574 demonstration trajectories and a shared pool of 3,092 offline rollout trajectories to train $\pi_{0.5}$ policies \citep{pi05}. Table~\ref{tab:realrobot} compares demonstration-only BC, BC with all rollouts, and ARS-based rollout selection with or without visual references. All demonstration chunks are retained. For ARS-based selection, we use the stage-dependent threshold rule in Section~\ref{sec:progress-guided-policy-learning}, with $B=8$ bins and $\alpha=30\%$. Configurations incorporating rollout data use stratified sampling with a fixed 1:1 ratio between demonstration and eligible rollout chunks. Each configuration uses one training seed and is evaluated over 200 complete-task trials.

The visual-reference condition provides three image--text examples of task-relevant states during ARS progress inference, without numerical progress annotations. Additional training, reference, and evaluation details are provided in Appendix~\ref{app:realrobot-details}.

\begin{table}[t]
\centering
\begin{minipage}[c]{0.50\linewidth}
    \centering
    \includegraphics[width=0.82\linewidth]{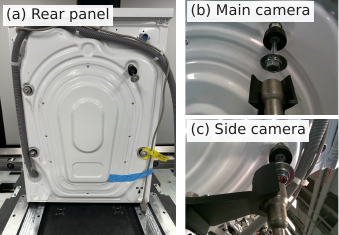}
\end{minipage}\hfill
\begin{minipage}[c]{0.47\linewidth}
    \captionof{table}{Real-world policy learning.}
    \label{tab:realrobot}
    \centering
    \normalsize
    \renewcommand{\arraystretch}{1.4}
    \setlength{\tabcolsep}{3pt}
    \begin{tabular*}{\linewidth}{@{\extracolsep{\fill}}lc@{}}
    \toprule
    Configuration & Success (\%) $\uparrow$ \\
    \midrule
    Demo + all rollouts BC & 50.5 \\
    Demo-only BC & 58.0 \\
    \arscolor{ARS} & 72.0 \\
    \arscolor{ARS} + visual reference & \textbf{81.5} \\
    \bottomrule
    \end{tabular*}
\end{minipage}
\par
\noindent\makebox[\linewidth][l]{%
\begin{minipage}{0.50\linewidth}
    \captionof{figure}{Screw-fastening setup.}
    \label{fig:realrobot-setup}
\end{minipage}}
\end{table}

Adding all rollouts reduces success from 58.0\% to 50.5\%, whereas ARS-selected rollouts raise it to 72.0\% (Table~\ref{tab:realrobot}), demonstrating the value of selective data incorporation. Visual references further raise observed success to 81.5\%, a total gain of 23.5 points over demonstration-only BC. Appendix~\ref{app:realrobot-outcome} reports additional outcome conditioning.

\section{Discussion}

ARS improves downstream policy performance while using fewer model queries and cumulative input tokens than frame-wise TOPReward. However, its iterative visual inspection and generative verification introduce additional inference overhead compared with single-forward-pass reward models such as Robometer. This tradeoff makes the current workflow particularly suited to offline data curation. Our progress estimates can also use later observations to resolve earlier ambiguities, and extending the workflow to causal online reward estimation remains an open direction. A promising direction is to distill ARS-generated supervision into lightweight reward models, combining the benefits of explicit verification with efficient deployment.

\subsection*{AI use statement}

We used generative AI tools to assist with research ideation and execution, including code development, literature retrieval and discovery, and drafting and polishing portions of the manuscript. We reviewed and approved all code, compiled and verified all experimental results, and revised and finalized the manuscript. We take full responsibility for the entire work, including all AI-assisted content and artifacts.

\subsection*{Reproducibility statement}

The main text describes the ARS inference workflow and experimental protocols. Additional details on agent prompts and tools, baseline implementations, policy training and evaluation, and ablation configurations are provided in the appendices. Code for ARS and semantic-mismatch benchmark generation is available at \url{https://github.com/midea-ai/ars}.

\bibliography{iclr2027_conference}
\bibliographystyle{iclr2027_conference}

\clearpage
\appendix
\section{Agent Prompts and Tool Set}
\label{app:agent-prompts-tools}

We summarize the main instructions and tools used by ARS. The prompt summaries retain the core reasoning and evidence requirements while omitting repetitive instructions and interface examples. The default workflow consists of event proposal, independent verification, and progress estimation. A subagent performs proposal generation; the primary agent performs verification and progress estimation in a separate reasoning context.

The proposal stage receives the full trajectory video when it fits within the configured input limit; otherwise, the agent inspects the trajectory in segments.

\subsection{Prompt summaries}

Table~\ref{tab:ars-prompt-summaries} summarizes the shared instructions, stage-specific instructions, and optional auxiliary context.

\begin{table}[H]
\caption{Main instructions used by ARS, summarized by component.}
\label{tab:ars-prompt-summaries}
\centering
\normalsize
\renewcommand{\arraystretch}{1.15}
\begin{tabular*}{\linewidth}{@{\extracolsep{\fill}}p{0.20\linewidth}p{0.75\linewidth}@{}}
\toprule
Component & Main instructions \\
\midrule
Shared instructions & Ground visual claims in inspected trajectory media. Use the task instruction to identify relevant entities and goal relations. Describe visible state changes and localize events using original frame indices. Request focused temporal windows and spatial crops when additional evidence could change the judgment. Express uncertainty through confidence scores and concise explanations. \\
\midrule
Event proposal & Construct an event timeline covering the entire trajectory. Organize intervals around meaningful changes in task state, including preparation, control acquisition, sustained manipulation, completion, recovery, and periods without progress. Associate each event with its frame interval, supporting visual evidence, and confidence. \\
\midrule
Independent verification & Treat proposed events as provisional claims. Acquire visual evidence during verification to examine each event's description, temporal boundaries, and confidence. Inspect neighboring frames to assess boundary placement. Revise, split, or merge events when necessary, and confirm each resulting event before proceeding. \\
\midrule
Progress estimation & For sequential tasks, infer subgoals and estimate total progress from their currently achieved contributions, including partial completion. Assign limited credit to useful preparation, localized increases to verified achievements, and gradual increases to continued advancement. Use plateaus for preserved states and decreases for deterioration. Later observations may verify an earlier event; assign credit to the supported event boundary. Preserve meaningful partial achievement in incomplete trajectories. \\
\midrule
Candidate-curve auditing & When an external progress curve is supplied, inspect its increases, decreases, and terminal values against the visual evidence. Retain supported estimates and revise unsupported intervals. Without a candidate curve, generate progress values for the complete trajectory. \\
\midrule
Optional auxiliary context & Use terminal outcome labels to constrain completion judgments while determining event timing from visual evidence. Use reference examples to interpret task states and goal relations. Apply reference progress values only when the corresponding state is visibly supported in the evaluated trajectory. \\
\bottomrule
\end{tabular*}
\end{table}

\clearpage
\subsection{Tool set}

The harness exposes tools according to the current stage. In Table~\ref{tab:ars-tool-set}, E denotes event proposal, V denotes verification, and P denotes progress estimation. Tools use the term \emph{observation} for an event record in the timeline.

\begin{table}[H]
\caption{Tools available in the default ARS workflow.}
\label{tab:ars-tool-set}
\centering
\normalsize
\renewcommand{\arraystretch}{1.15}
\begin{tabular*}{\linewidth}{@{\extracolsep{\fill}}p{0.32\linewidth}cp{0.52\linewidth}@{}}
\toprule
Tool & Stage & Function \\
\midrule
\texttt{read\_robot\_instruction} & E, V, P & Retrieve the task instruction and any explicitly supplied task clarification. \\
\midrule
\texttt{read\_metadata} & E, V, P & Retrieve technical video metadata, including frame count, frame dimensions, presentation frame rate, and frame-indexing conventions. This tool provides no additional task annotations or simulator state. \\
\midrule
\texttt{inspect\_visual} & E, V, P & Retrieve a selected frame interval as a video clip or image tile, with configurable temporal sampling and an optional spatial crop. Return the media with its evidence identifier and frame indices. \\
\midrule
\texttt{read\_observations} & E, V, P & Retrieve the current event timeline and coverage diagnostics. \\
\midrule
\texttt{add\_observations}\newline\texttt{revise\_observations} & E, V & Create or revise event intervals, descriptions, evidence links, and confidence scores. \\
\midrule
\texttt{merge\_observations} & V & Combine events that describe one continuous phase. \\
\midrule
\texttt{confirm\_observations} & V & Confirm an event using evidence acquired during verification. \\
\midrule
\texttt{read\_progress}\newline\texttt{summarize\_progress} & P & Inspect candidate and current progress values, curve statistics, and jump locations and magnitudes. \\
\midrule
\texttt{edit\_progress\_segments} & P & Generate or revise local intervals by interpolating between specified endpoint values, supporting increases, plateaus, and decreases. \\
\midrule
\texttt{calibrate\_progress\_\allowbreak curve} & P & Apply a global affine calibration using two frame--value anchors when the curve's relative trend is supported by the evidence. \\
\midrule
\texttt{validate\_progress} & P & Check frame assignments, numerical bounds, and configured curve constraints. \\
\midrule
\texttt{write\_summary} & P & Record the final judgment, supporting evidence, progress edits, and need for external review. \\
\midrule
\texttt{complete\_stage} & E, V, P & Check stage requirements and advance the workflow; completion of the final stage exports the results. \\
\bottomrule
\end{tabular*}
\end{table}

The agent represents progress using integer points in $[0,1000]$, corresponding to a normalized progress scale of $[0,1]$.

Observation creation, revision, and merging, as well as progress editing and calibration, use a shared confidence scale from 1 (very uncertain) to 5 (very high confidence). Scores reflect the clarity of the supporting evidence. Progress edits and calibration also include an evidence-grounded explanation.

Cited evidence must cover each event's recorded interval, and the event collection must cover the entire trajectory. Verification considers context before and after each interval when available. Revisions reset confirmation status.

To finalize progress estimates, the agent validates the curve, writes the summary, and completes the stage, in that order. Blocked tool calls report unmet requirements and corrective actions.

\section{ARS Reasoning Trace on a LIBERO Trajectory}
\label{app:ars-reasoning-trace}

We examine the ARS reasoning trace for the partially successful LIBERO trajectory in Figure~\ref{fig:ars-timeline}. The trajectory contains 520 frames, indexed from 0 to 519, under the instruction ``turn on the stove and put the moka pot on it.'' This case illustrates how verification changes event boundaries, how tool feedback guides retries, and where errors remain after verification.

\paragraph{Subagent inspection and event proposal.}
The subagent receives the complete trajectory video and makes nine additional \texttt{inspect\_visual} calls to examine knob manipulation, moka pot grasping and placement, the final state, and transitions between these phases. It then proposes the eight events listed in Table~\ref{tab:ars-trace-events}.

\begin{table}[H]
\caption{Subagent-proposed event timeline for the LIBERO trajectory in Figure~\ref{fig:ars-timeline}, before primary-agent verification.}
\label{tab:ars-trace-events}
\centering
\normalsize
\renewcommand{\arraystretch}{1.15}
\begin{tabular}{lc}
\toprule
Proposed event & Frame interval \\
\midrule
Initial state and approach to the stove knob & $[0,94]$ \\
Knob grasping and turning & $[95,127]$ \\
Burner switching off and withdrawal & $[128,180]$ \\
Repositioning and approach to the moka pot & $[181,269]$ \\
Moka pot grasp acquisition & $[270,295]$ \\
Lifting and transporting the moka pot & $[296,339]$ \\
Placement on the stove burner & $[340,370]$ \\
Final state: pot on the burner, burner off & $[371,519]$ \\
\bottomrule
\end{tabular}
\end{table}

\paragraph{Primary-agent verification and boundary refinement.}
The primary agent receives this timeline in a separate reasoning context and makes eight additional \texttt{inspect\_visual} calls. The main revision concerns the boundary between \emph{Initial state and approach} and \emph{Knob grasping and turning}. After inspecting frames $[0,100]$, the primary agent judges that knob grasping begins substantially earlier than the proposed frame 95. It shortens the approach event to $[0,45]$ and moves the start of knob manipulation to frame 46.

Manual inspection identifies frame 56 as the onset of actual control over the knob. Relative to this annotation, verification reduces the boundary error from 39 frames late to 10 frames early. The revised boundary is therefore substantially closer to the annotated transition, although it still assigns control acquisition prematurely. This case illustrates both the benefit of independent verification and the underlying VLM's remaining difficulty in precisely localizing subtle manipulation events.

\paragraph{Blocked calls and corrective retries.}
The initial boundary-revision call is blocked because the cited clip $[90,140]$ does not cover the expanded event interval $[46,126]$. The agent adds an already inspected clip covering $[0,100]$ and retries successfully. This illustrates how evidence-coverage feedback guides correction of tool calls.

The primary agent subsequently uses the verified timeline to assign progress, capturing the regression when the stove is unintentionally switched off and crediting the later manipulation and placement of the moka pot.

\section{Implementation Details of Compared Methods}
\label{app:compared-methods}

We evaluate the compared methods using the implementations and input protocols described below:
\begin{itemize}[leftmargin=*]
    \item \textbf{\lrmcolor{LRM}} \citep{lrm}: we use its Absolute Progress model, which estimates absolute task progress for each frame and outputs one of 11 discrete values, $\{0.0, 0.1, \ldots, 1.0\}$.
    \item \textbf{\rtwovlmcolor{R\textsuperscript{2}VLM}} \citep{r2vlm} recurrently processes video snippets, carrying forward a CoT that tracks subtask completion to estimate current task progress. We use its best available checkpoint.
    \item \textbf{\solercolor{SOLE-R1}} \citep{soler1} performs per-timestep spatiotemporal CoT reasoning to estimate dense task progress.
    \item \textbf{\robometercolor{Robometer}} \citep{robometer} was trained on sequences of 8 sampled video frames to predict per-frame task progress. We also tested sliding-window inference and larger input frame counts on long-horizon tasks, but observed no improvement over the original 8-frame setting. We therefore retain 8 sampled frames per trajectory.
    \item \textbf{\toprewardcolor{TOPReward}} \citep{topreward} is a reward-training-free method that scores video prefixes conditioned on the task instruction using the log-probability of the \texttt{True} token in a task-completion query. Following the model choices in the original work, we evaluate TOPReward with Qwen3-VL-8B-Instruct and Qwen3-VL-32B-Instruct \citep{qwen3vl}. At each timestep, we run inference using the task instruction and all video frames up to and including the current frame.
\end{itemize}

SOLE-R1 uses both the main and wrist cameras as inputs, while all other methods, including ARS, use only the main camera. We provide all trajectory frames as input to LRM, R\textsuperscript{2}VLM, SOLE-R1, and TOPReward. For Robometer, we follow the frame-sampling protocol specified in its paper, including both the first and last frames. We linearly interpolate between adjacent predictions to construct dense progress curves. Metrics in Table~\ref{tab:mismatch-main} are computed using each method's native model outputs, excluding interpolated values.

\paragraph{Output normalization.}
\label{app:output-normalization}
Let $p_t^{\mathrm{raw}}$ denote a method's raw output and $p_t$ its evaluated progress. We use the following fixed output conventions; they specify numerical scales without assuming a common calibration across methods.
\begin{itemize}[leftmargin=*]
    \item \textbf{\lrmcolor{LRM}} and \textbf{\robometercolor{Robometer}} output progress in $[0,1]$ and require no rescaling.
    \item \textbf{\rtwovlmcolor{R\textsuperscript{2}VLM}} outputs values in $[0,100]$; we use $p_t=p_t^{\mathrm{raw}}/100$.
    \item \textbf{\solercolor{SOLE-R1}} outputs signed values in $[-100,100]$. The main results use $p_t=p_t^{\mathrm{raw}}/100\in[-1,1]$, preserving its protocol-defined initial zero. We additionally report the affine mapping $p_t^{\mathrm{aff}}=(p_t^{\mathrm{raw}}+100)/200=(p_t+1)/2\in[0,1]$ as a diagnostic.
    \item \textbf{\toprewardcolor{TOPReward}} outputs the log-probability of the \texttt{True} token, $p_t^{\mathrm{raw}}\in(-\infty,0]$. Following its original reward definition, our primary output convention uses the fixed linear normalization $p_t^{\mathrm{lin}}=(\max(p_t^{\mathrm{raw}},-30)+30)/30$. The minimum raw output observed across all our experiments was $-28.5$, so no output was clipped by the $-30$ lower bound. We additionally evaluate the corresponding token probability, $p_t^{\mathrm{prob}}=\exp(p_t^{\mathrm{raw}})$; Appendix~\ref{app:semantic-mismatch-details} compares the two conventions. Exponentiation compresses absolute increments in low-probability regions, making them more susceptible to filtering by an absolute tolerance.
    \item \textbf{\arscolor{ARS}} outputs integer progress points in $[0,1000]$; dividing by 1000 gives progress in $[0,1]$.
\end{itemize}

For the TOPReward curve in Figure~\ref{fig:ars-timeline},
we use Qwen3-VL-32B-Instruct and follow the original paper's
within-trajectory min--max
normalization \citep{topreward}, using
$p_t^{\mathrm{vis}}=(p_t^{\mathrm{raw}}-p_{\min}^{\mathrm{raw}})/(p_{\max}^{\mathrm{raw}}-p_{\min}^{\mathrm{raw}})$.
The raw log-probabilities over this trajectory have a minimum
of $-12.9$ and a maximum of
$-0.0014$. This normalization applies only to this visualization.

\section{Additional Details on the Semantic-Mismatch Benchmark}
\label{app:semantic-mismatch-details}

\subsection{Training-data exposure to ManiSkill}
\label{app:maniskill-exposure}

We summarize the training-data exposure of the evaluated reward models to ManiSkill based on the original papers and model documentation. $\mathbb{E}$ indicates documented use of ManiSkill data during training, including reward learning and other supervised fine-tuning. N indicates that no evidence of such use was found. These labels describe benchmark-family exposure and do not establish overlap with our specific evaluation trajectories. N does not rule out exposure during training of the underlying VLM.

\begin{table}[H]
\caption{Training-data exposure of reward models to ManiSkill.}
\label{tab:maniskill-training-exposure}
\centering
\normalsize
\renewcommand{\arraystretch}{1.15}
\begin{tabular*}{\linewidth}{@{\extracolsep{\fill}}p{0.18\linewidth}cp{0.64\linewidth}@{}}
\toprule
Method & Exposure & Evidence \\
\midrule
\lrmcolor{LRM} & N & ManiSkill is described as zero-shot to the reward models \citep{lrm}. \\
\rtwovlmcolor{R\textsuperscript{2}VLM} & N & No ManiSkill training is reported for the evaluated checkpoint \citep{r2vlm}. \\
\solercolor{SOLE-R1} & N & No ManiSkill training source is identified in the published training mixture \citep{soler1}. \\
\robometercolor{Robometer} & $\mathbb{E}$ & RBM-1M includes FAILSafe trajectories collected in ManiSkill \citep{robometer}. \\
\toprewardcolor{TOPReward} & N & No ManiSkill training is reported in the Qwen model cards; no additional reward-model training is performed \citep{topreward}. \\
\arscolor{ARS} & N & No ManiSkill training is reported in the Qwen model cards; no additional reward-model training is performed. \\
\bottomrule
\end{tabular*}
\par\smallskip
{\normalsize\raggedright Qwen model cards: \href{https://huggingface.co/Qwen/Qwen3-VL-8B-Instruct}{Qwen3-VL-8B-Instruct}, \href{https://huggingface.co/Qwen/Qwen3-VL-32B-Instruct}{Qwen3-VL-32B-Instruct}, \href{https://huggingface.co/Qwen/Qwen3.6-27B}{Qwen3.6-27B}, and \href{https://huggingface.co/Qwen/Qwen3.5-4B}{Qwen3.5-4B}.\par}
\end{table}

\subsection{False Progress Suppression Ratio (FPSR)}
\label{app:fpsr}

\paragraph{Definition.}
For mismatch trajectory $i$, let $0=t_0<\cdots<t_K=T_i$ be the frame indices of its native predictions $p_{t_0},\ldots,p_{t_K}$, where $T_i$ is the terminal frame index. Omitting the trajectory index on the prediction times and values for readability, we define, for a tolerance $\eta\geq0$,
\begin{equation}
    \mathrm{FPSR}_i(\eta)
    = 1-
    \frac{
        \sum_{k=0}^{K-1}(t_{k+1}-t_k)\,
        \mathbbm{1}[p_{t_{k+1}}-p_{t_k}>\eta]
    }{T_i}.
    \label{eq:fpsr}
\end{equation}
The dataset score is the unweighted mean over mismatch trajectories. We compute the metric from adjacent native predictions without interpolation. Its value remains sensitive to the temporal spacing of these predictions.

\paragraph{Expert progress coverage ratio (EPCR).}
Increasing $\eta$ can suppress useful progress estimates as well as spurious increases. We measure the fraction of each correctly instructed expert trajectory's duration covered by intervals whose predicted progress increase exceeds $\eta$, and report its unweighted mean across trajectories as $\mathrm{EPCR}(\eta)$. This uses the fraction in Equation~\ref{eq:fpsr}, without the leading $1-$, evaluated under correct instructions. The expert trajectories in this benchmark advance the correctly instructed task at every recorded step, making EPCR informative about the availability of the intended dense signal. A predictor that is flat on both instruction groups has FPSR equal to one but EPCR equal to zero.

\paragraph{Tolerance sensitivity.}
Table~\ref{tab:fpsr-sensitivity} reports FPSR together with $\mathrm{EPCR}$ for $\eta\in\{0,10^{-4},10^{-3},0.005,0.01\}$. The main metric uses $\eta=0$. Only increments exceeding $\eta$ are counted as progress increases. For integer-valued outputs, we compare raw integer differences against the appropriately scaled threshold to avoid floating-point subtraction errors at the boundary. The output conventions are specified in Appendix~\ref{app:output-normalization}.

\begin{table}[t]
\caption{Tolerance sensitivity on the semantic-mismatch benchmark. Each entry is FPSR / EPCR, both on a $[0,1]$ scale, using native predictions. TOPReward-8B and -32B use Qwen3-VL. ARS and Audit use Qwen3.6-27B. TOPReward probability outputs are reported in the diagnostics section.}
\label{tab:fpsr-sensitivity}
\centering
\normalsize
\setlength{\tabcolsep}{2pt}
\renewcommand{\arraystretch}{1.12}
\begin{tabular*}{\linewidth}{@{\extracolsep{\fill}}lccccc@{}}
\toprule
Method & $0$ & $10^{-4}$ & $10^{-3}$ & $0.005$ & $0.01$ \\
\midrule
\lrmcolor{LRM} & $0.893\,/\,0.159$ & $0.893\,/\,0.159$ & $0.893\,/\,0.159$ & $0.893\,/\,0.159$ & $0.893\,/\,0.159$ \\
\rtwovlmcolor{R\textsuperscript{2}VLM} & $0.184\,/\,0.791$ & $0.184\,/\,0.791$ & $0.184\,/\,0.791$ & $0.184\,/\,0.791$ & $0.190\,/\,0.786$ \\
\solercolor{SOLE-R1}, $[-1,1]$ & $0.361\,/\,0.731$ & $0.361\,/\,0.731$ & $0.361\,/\,0.731$ & $0.361\,/\,0.731$ & $0.371\,/\,0.729$ \\
\robometercolor{Robometer} & $0.240\,/\,0.978$ & $0.242\,/\,0.977$ & $0.246\,/\,0.976$ & $0.273\,/\,0.971$ & $0.306\,/\,0.965$ \\
\toprewardcolor{TOPReward}-8B, linear & $0.368\,/\,0.656$ & $0.383\,/\,0.642$ & $0.390\,/\,0.622$ & $0.452\,/\,0.560$ & $0.508\,/\,0.509$ \\
\toprewardcolor{TOPReward}-32B, linear & $0.383\,/\,0.723$ & $0.412\,/\,0.698$ & $0.447\,/\,0.660$ & $0.585\,/\,0.526$ & $0.693\,/\,0.406$ \\
\arscolor{ARS} Audit & $0.915\,/\,0.899$ & $0.915\,/\,0.899$ & $0.919\,/\,0.885$ & $0.942\,/\,0.691$ & $0.960\,/\,0.578$ \\
\arscolor{ARS}-27B & $0.918\,/\,0.908$ & $0.918\,/\,0.908$ & $0.921\,/\,0.902$ & $0.940\,/\,0.717$ & $0.960\,/\,0.588$ \\
\midrule
\multicolumn{6}{l}{\textit{Additional output-scale diagnostics}} \\
\solercolor{SOLE-R1}, $[0,1]$ & $0.361\,/\,0.731$ & $0.361\,/\,0.731$ & $0.361\,/\,0.731$ & $0.371\,/\,0.729$ & $0.500\,/\,0.641$ \\
\toprewardcolor{TOPReward}-8B, prob. & $0.368\,/\,0.656$ & $0.853\,/\,0.354$ & $0.947\,/\,0.277$ & $0.976\,/\,0.213$ & $0.982\,/\,0.183$ \\
\toprewardcolor{TOPReward}-32B, prob. & $0.383\,/\,0.723$ & $0.583\,/\,0.583$ & $0.792\,/\,0.490$ & $0.927\,/\,0.371$ & $0.955\,/\,0.296$ \\
\bottomrule
\end{tabular*}
\end{table}

Table~\ref{tab:fpsr-sensitivity} shows that ARS combines high false-progress suppression with high expert progress coverage. For $0\leq\eta\leq10^{-3}$, its FPSR remains 0.918--0.921 and $\mathrm{EPCR}$ remains 0.902--0.908. By comparison, LRM achieves FPSR of 0.893 but covers only 0.159 of expert-trajectory duration.

Increasing the tolerance can raise FPSR while removing useful progress signals. For TOPReward-8B probability outputs, increasing $\eta$ from zero to $10^{-4}$ raises FPSR from 0.368 to 0.853 while reducing $\mathrm{EPCR}$ from 0.656 to 0.354. Even ARS loses substantial coverage at larger tolerances, with $\mathrm{EPCR}$ falling to 0.588 at $\eta=0.01$. These gains in FPSR therefore require joint interpretation with the remaining expert progress coverage.

\subsection{Correct--Mismatch Progress Gap (CMPG)}
\label{app:cmpg}

\paragraph{Definition and motivation.}
Robometer \citep{robometer} evaluates the difference in terminal rewards between successful and failed trajectories of the same task. Our semantic-mismatch benchmark imposes a stronger condition on the failure cases: the robot never approaches the object specified by the mismatch instruction or advances the corresponding task. Progress should therefore remain low throughout each mismatch trajectory. We accordingly replace terminal mismatch progress with the maximum predicted progress over the trajectory, so that transient false progress is penalized even if the prediction later decreases.

For each of the $N=500$ trajectory pairs, let $p_{i,t}^{+}$ and $p_{i,t}^{-}$ denote progress under the correct and mismatch instructions, respectively. Let $T_i$ denote the terminal timestep and $\mathcal T_i^{-}$ the timesteps with native model predictions under the mismatch instruction. We define
\begin{equation}
    \mathrm{CMPG}
    =
    \frac{1}{N}\sum_{i=1}^{N} p_{i,T_i}^{+}
    -
    \frac{1}{N}\sum_{i=1}^{N}
    \max_{t\in\mathcal T_i^{-}} p_{i,t}^{-}.
    \label{eq:cmpg}
\end{equation}
Each trajectory contributes equally, irrespective of its length or number of predictions. Higher CMPG indicates greater separation between correct terminal progress and peak false progress.

CMPG complements VOC and FPSR by measuring numerical separation between correct terminal progress and peak mismatch progress. For example, a model that predicts a perfectly increasing progress curve from 0 to 1 under the correct instruction, but a constant progress of 0.99 under the mismatch instruction, achieves perfect VOC and FPSR. Nevertheless, it assigns nearly complete progress to an unrelated task, which is reflected in its low CMPG of 0.01. Although CMPG is sensitive to the output scale, we retain it to capture this complementary aspect of task alignment and interpret it jointly with VOC and FPSR under the normalization conventions in Appendix~\ref{app:output-normalization}.

\paragraph{Analysis under different output conventions.}
For SOLE-R1, the main $[-1,1]$ convention preserves its protocol-defined initial zero. Mapping the same predictions to $[0,1]$ via $p_t^{\mathrm{aff}}=(p_t+1)/2$ halves CMPG from 0.266 to 0.133: the common offset cancels, while the gap scales by one half. VOC and zero-threshold FPSR and EPCR remain unchanged.

For TOPReward, replacing linearly normalized log-probabilities with token probabilities increases CMPG from 0.190 to 0.685 with Qwen3-VL-8B and from 0.129 to 0.756 with Qwen3-VL-32B. This transformation preserves VOC and zero-threshold FPSR and EPCR; the larger CMPG values reflect the nonlinear change in output scale. We retain linearly normalized log-probabilities as our primary output convention, following TOPReward's original reward definition \citep{topreward}. In Table~\ref{tab:mismatch-main}, we report the probability-based CMPG values to present the more favorable results for TOPReward among the two evaluated conventions. Both values remain below ARS's CMPG of 0.904. Table~\ref{tab:fpsr-sensitivity} further shows that, across the tested positive tolerances, probability outputs exhibit larger FPSR increases and EPCR reductions than the linear convention.

\section{Additional Details on LIBERO-Long}
\label{app:libero-long-details}

\subsection{Training-data exposure to LIBERO}
\label{app:libero-exposure}

Table~\ref{tab:libero-training-exposure} summarizes training-data exposure to LIBERO based on the original papers and model documentation. $\mathbb{E}$ indicates documented use of LIBERO data during training, and N indicates that no evidence of such use was found. The labels describe benchmark-family exposure without establishing overlap with our specific policy-training trajectories. N does not rule out exposure during training of the underlying VLM.

\begin{table}[H]
\caption{Training-data exposure of reward models to LIBERO.}
\label{tab:libero-training-exposure}
\centering
\normalsize
\renewcommand{\arraystretch}{1.15}
\begin{tabular*}{\linewidth}{@{\extracolsep{\fill}}p{0.18\linewidth}cp{0.64\linewidth}@{}}
\toprule
Method & Exposure & Evidence \\
\midrule
\lrmcolor{LRM} & $\mathbb{E}$ & LIBERO is an explicit source in the training mixture \citep{lrm}. \\
\rtwovlmcolor{R\textsuperscript{2}VLM} & N & No LIBERO training is reported for the evaluated checkpoint \citep{r2vlm}. \\
\solercolor{SOLE-R1} & $\mathbb{E}$ & Supervised fine-tuning includes 504,544 \texttt{ecot\_libero\_all} examples \citep{soler1}. \\
\robometercolor{Robometer} & $\mathbb{E}$ & RBM-1M includes LIBERO-\{Long, Object, Spatial, Goal\} and generated failures \citep{robometer}. \\
\toprewardcolor{TOPReward} & N & No LIBERO training is reported in the Qwen model cards; no additional reward-model training is performed \citep{topreward}. \\
\arscolor{ARS} & N & No LIBERO training is reported in the Qwen model cards; no additional reward-model training is performed. \\
\bottomrule
\end{tabular*}
\end{table}

\subsection{Task instruction clarifications}
\label{app:task-instruction-clarifications}

For four of the LIBERO-Long tasks, we append brief clarifications to the original task instructions to clarify object identities and disambiguate spatial references. We provide identical clarifications to all reward systems, including ARS and every compared baseline, ensuring that all VLM-based methods receive the same additional semantic information.

For the three basket-placement tasks, the object names in the original instructions can be difficult to match to the visible entities. At the relatively low default resolution of $256 \times 256$ pixels, packaging text and graphics can be difficult to discern. Without background information about the objects, even human viewers can struggle to establish this correspondence. We therefore append short descriptions of object color, shape, and packaging.

For the mug-placement task, the original instruction specifies ``left'' and ``right'' without identifying the reference frame. We clarify that these directions are robot-relative and provide the corresponding image directions. Table~\ref{tab:libero-instruction-clarifications} lists the original instructions and the appended clarifications.

\begin{table}[ht]
\caption{Task instruction clarifications used on LIBERO-Long. The same additions are provided to ARS and every compared baseline. Each addition is appended to the original instruction under the heading ``Task-specific hints:''.}
\label{tab:libero-instruction-clarifications}
\centering
\normalsize
\renewcommand{\arraystretch}{1.2}
\begin{tabular*}{\linewidth}{
    @{\extracolsep{\fill}}
    p{0.43\linewidth}
    p{0.53\linewidth}
    @{}
}
\toprule
\textbf{Original instruction} & \textbf{Appended clarification} \\
\midrule
put both the alphabet soup and the tomato sauce in the basket
&
alphabet soup: blue-and-orange can with alphabet-letter soup graphics.
\newline
tomato sauce: red/green/white can with tomato pictures.
\\
\midrule
put both the alphabet soup and the cream cheese box in the basket
&
alphabet soup: blue-and-orange can with alphabet-letter soup graphics.
\newline
cream cheese: white rectangular box with a large purple-and-blue label.
\\
\midrule
put both the cream cheese box and the butter in the basket
&
cream cheese: white rectangular box with a large purple-and-blue label.
\newline
butter: red-orange rectangular box with a blue-and-white panel.
\\
\midrule
put the white mug on the left plate and put the yellow and white mug on the right plate
&
Plate sides are robot-relative: white mug goes to robot-left (image right); yellow-and-white mug goes to robot-right (image left).
\\
\bottomrule
\end{tabular*}
\end{table}

\subsection{Hand-designed privileged progress}
\label{app:privileged-progress}

We construct a progress signal from simulator-provided goal predicates and manually specified scoring rules. All scoring parameters were fixed before downstream policy training. For placement goals, the signal combines reaching the designated object, lifting it, and reducing its distance to the target. For articulated goals, such as opening or closing an object, progress is based on changes in joint distance to the target range. This configuration uses privileged task information and simulator state during offline data selection and follows the same chunk-selection, policy-training, and evaluation protocol as the other progress-based configurations.

For example, the task ``put both the alphabet soup and the tomato sauce in the basket'' contains two object-placement goals. For each object $i$, we compute normalized reaching, lifting, and placement scores $R_i,L_i,D_i\in[0,1]$. Reaching depends on the gripper--object distance, while lifting measures the object's height increase relative to reset, modulated by gripper proximity. Placement measures the reduction in distance to the basket's containment region relative to reset. This distance is computed in three dimensions, with a 1\,cm downward tolerance at the region's lower boundary. Before the corresponding containment predicate is satisfied, its score is

\begin{equation}
    u_i=\min\!\left(
        0.98,\,
        \max\!\left\{
            0.15R_i+0.35L_i+0.50D_i,\;
            0.35L_i+0.65D_i
        \right\}
    \right).
    \label{eq:privileged-placement-score}
\end{equation}

For these initially unsatisfied goals, we set $q_i=1$ when the environment's corresponding containment predicate is satisfied and $q_i=u_i$ otherwise. Overall progress is $p_t=(q_1+q_2)/2$, with the environment's task-success check explicitly setting $p_t=1$. We apply the same selection rule as in the other progress-based configurations, retaining chunks with $p_{t+H}-p_t>0$, where $H=10$ for LIBERO-Long.

\subsection{Policy training and evaluation}
\label{app:bc-training-evaluation}

\paragraph{Protocol for comparing reward systems.}
Our simulation experiments evaluate the utility of different reward systems for downstream policy learning under a shared protocol. Following prior work on weighted behavior cloning \citep{vip,gvl,sarm,warp-rm,arm}, we use predicted progress changes to determine the contribution of each action chunk to policy training. All reward systems operate on the same fixed rollout dataset and share the policy initialization, training budget, and evaluation procedure. This setup provides a direct connection between progress estimates and policy supervision without introducing an additional learned critic.

We also considered continuous weighting based on progress increments. However, different reward systems produce progress curves with different temporal profiles and distributions of increment magnitudes, making it difficult to choose shared weighting hyperparameters that yield comparable weighting behavior across methods. Tuning these hyperparameters separately would introduce additional optimization choices into the comparison. We therefore retain chunks with $\Delta_t>0$ and assign them equal weight. Although this is a coarse criterion that discards progress magnitude, it provides a common comparison protocol with no additional selection hyperparameters and reduces dependence on the numerical calibration of progress estimates.

The zero-threshold rule also retains slow but potentially necessary transitions whenever their predicted progress change is positive. Selecting only high-gain chunks within each progress stage may reduce coverage of these actions. Adding unfiltered demonstrations can mitigate this issue but introduces another source of supervision, making downstream performance a less direct measure of rollout selection quality. We therefore use a zero threshold for the simulation comparison and stage-dependent thresholds for the real-robot experiments, where all demonstration chunks are retained for training.

\paragraph{Training and evaluation schedule.}
We train all policies for 30k gradient steps with three independent training seeds, using the same global batch size and optimization schedule. Following the LIBERO evaluation setup of OpenVLA-OFT \citep{openvla-oft}, we evaluate all methods every 2k steps on all ten LIBERO-Long tasks, increasing the budget to 100 rollouts per task (1{,}000 per checkpoint). Each rollout allows up to 520 control steps, with the same benchmark-provided initial states and evaluation seed across methods. Following the best-checkpoint reporting convention in LIBERO and robomimic \citep{libero,robomimic}, we select the checkpoint with the highest mean success rate across the ten tasks for each training seed and report the mean and sample standard deviation of these peak rates across the three seeds. Checkpoint selection and reporting use the same benchmark evaluation set.

Different reward systems retain different subsets of the shared rollout dataset, producing policy-training datasets that vary in both size and composition. We therefore fix the number of gradient updates rather than the number of epochs to maintain a matched optimization budget across methods; methods retaining fewer chunks revisit their training samples more frequently. Under this training configuration, we observe that evaluation success rates plateau for all methods, with no sustained improvement toward the end of the 30k-step training budget. As a diagnostic of performance at a fixed training step, we additionally report results at the final 30k-step checkpoint for the main method comparison (Table~\ref{tab:libero-final-checkpoint}).

\paragraph{Results at the final checkpoint.}
ARS achieves 51.6\% success at the final checkpoint, exceeding Vanilla BC by 11.5 percentage points and the strongest baseline, Robometer, by 9.5 points.

\begin{table}[ht]
\caption{Policy-learning results on LIBERO-Long at the final 30k-step checkpoint. Entries report mean success rate $\pm$ standard deviation over the same three training runs per method as in Table~\ref{tab:libero}.}
\label{tab:libero-final-checkpoint}
\centering
\normalsize
\begin{tabular}{lc}
\toprule
Method & Success (\%) $\uparrow$ \\
\midrule
Vanilla BC & 40.1 $\pm$ 1.8 \\
\midrule
\lrmcolor{LRM} & 36.8 $\pm$ 1.5 \\
\rtwovlmcolor{R\textsuperscript{2}VLM} & 34.1 $\pm$ 1.1 \\
\solercolor{SOLE-R1} & 40.4 $\pm$ 2.8 \\
\robometercolor{Robometer} & 42.1 $\pm$ 2.3 \\
\midrule
\toprewardcolor{TOPReward} + Qwen3-VL-8B & 37.9 $\pm$ 2.5 \\
\toprewardcolor{TOPReward} + Qwen3-VL-32B & 40.5 $\pm$ 2.9 \\
\arscolor{ARS} + Qwen3.6-27B & \textbf{51.6} $\pm$ 1.4 \\
\bottomrule
\end{tabular}
\end{table}

\subsection{Progress-curve diagnostics}
\label{app:progress-diagnostics}

\paragraph{Zero progress increments.}
In Table~\ref{tab:libero}, LRM and R\textsuperscript{2}VLM retain only 13.7\% and 13.6\% of the chunks, respectively, and yield the two lowest downstream success rates, 38.6\% and 41.5\%. For LRM, the 11-level absolute-progress output produces zero progress differences for 76.7\% of candidate chunks. R\textsuperscript{2}VLM assigns $\Delta_t=0$ to 79.5\% of chunks because it often repeats an early progress estimate across subsequent frames, leaving little temporal information for chunk selection.

\paragraph{Local progress under outcome conditioning.}
We examine adjacent-frame progress changes on the 3,098 trajectories where the conditioned and unconditioned runs agree on terminal outcome. With outcome conditioning, non-increasing intervals on success-predicted trajectories remain at 21.45\% (21.70\% without conditioning), while failure-predicted trajectories retain 23.38\% increasing intervals (27.03\% without conditioning), micro-averaged within each group. These patterns are consistent with the terminal label serving as an endpoint anchor while intermediate progress remains guided by visual evidence, rather than collapsing into a uniformly increasing ramp for success or a flat zero curve for failure.

\section{Additional Ablation Details}
\label{app:ablation-details}

\subsection{Base-model evaluation and compatibility}
\label{app:base-model-ablations}

\subsubsection{Additional base-model evaluation for TOPReward}
\label{app:topreward-base-model}

Following the authors' public inference implementation \citep{topreward}, we evaluate TOPReward with Qwen3-VL-8B-Instruct and Qwen3-VL-32B-Instruct and obtain reasonable results on both evaluations reported in Tables~\ref{tab:mismatch-main} and~\ref{tab:libero}. To match ARS's default base model, we additionally evaluate TOPReward with Qwen3.6-27B. Starting from the reference settings, we test alternative prompt and chat-template formats, as well as enabling and disabling thinking mode. These attempts do not yield competitive results in our setup. The reported Qwen3.6-27B configuration achieves 0.452 VOC, 0.471 FPSR, and $-0.049$ CMPG under the linear log-probability convention described in Appendix~\ref{app:output-normalization}.

We hypothesize that this difficulty may partly reflect TOPReward's sensitivity to prompt formatting, as documented by the original authors \citep{topreward}. Qwen3.6-27B uses a different chat template from Qwen3-VL-Instruct, including different handling of thinking prefixes \citep{qwen3vl,qwen36}, which may affect completion-token probabilities. However, our experiments do not isolate these differences as the cause of the performance degradation. We retain the original Qwen3-VL model family in the main comparisons and report the Qwen3.6-27B adaptation here.

\subsubsection{Base-model execution requirements}
\label{app:base-model-execution}

To compare ARS with Robometer using the same base model, we attempted to run ARS with Qwen3-VL-4B-Instruct, the base model from which Robometer was trained \citep{robometer}. Despite its visual understanding capabilities, the model failed to complete the ARS workflow in any of our runs. Failures involved tool use and compliance with harness constraints. We then evaluated Qwen3.5-4B, a model of the same nominal parameter size from the Qwen family, which completed all 1,000 semantic-mismatch evaluations. We therefore use Qwen3.5-4B for the comparison with Robometer, matching the parameter scale while using a different base model.

\subsubsection{Small-model execution on long-horizon trajectories}
\label{app:small-model-long-horizon}

On LIBERO-Long, Qwen3.5-4B produced valid ARS outputs for 4,063 of 4,096 trajectories, an inference completion rate of 99.19\%. The remaining 33 runs were marked as failures after extended attempts, including cases of repeated violations of harness constraints. These observations indicate greater difficulty sustaining reliable agent execution on long-horizon trajectories, even when the same model completes the simpler semantic-mismatch benchmark.

\subsection{Workflow ablations: implementation details and additional experiments}
\label{app:workflow-ablations}

\subsubsection{ARS workflow variants}

We compare full ARS with three ablated workflows. The first omits both visual inspection and explicit event verification, relying on the full video supplied initially. The second retains visual inspection but omits explicit event verification. The third retains the backbone, tools, and verification procedure of full ARS but performs proposal generation and verification within a single reasoning context. Progress estimation continues in that same context. Full ARS instead generates proposals in one context and performs verification and progress estimation in a separate context. Across variants, we use Qwen3.6-27B with the same inference settings and progress-scoring criteria, and provide all video frames at their original resolution. A workflow completes when the agent requests completion and passes the applicable harness checks.

Actual query and token counts vary with workflow execution and the agents' stopping decisions. A query is one model scoring or generation request. We sum usage across all agents and report the mean per trajectory over all 4,096 trajectories. Input and output tokens are taken from the reported prompt and completion usage, respectively. Input counts include repeated context and visual tokens; completion counts already include reasoning tokens.

During verification and progress estimation, the primary agent averages 41.4K input tokens per query, 28.1\% below the single-context control's average of 57.6K across the complete workflow (Table~\ref{tab:ars-agent-usage}). This is consistent with the benefit of context separation described in Section~\ref{sec:cross-agent-verification}: retaining event proposals without carrying over the subagent's full interaction history.

\begin{table}[ht]
\caption{Per-agent inference usage of full ARS on LIBERO-Long, with the single-context control for comparison. Queries and token counts are means per trajectory. Input per query is aggregate input tokens divided by model queries.}
\label{tab:ars-agent-usage}
\begin{center}
\normalsize
\setlength{\tabcolsep}{4pt}
\begin{tabular*}{\linewidth}{@{\extracolsep{\fill}}lrrrr@{}}
\toprule
Configuration & Queries & Input (K) & Output (K) & Input / query (K) \\
\midrule
Full \arscolor{ARS}: subagent & 5.8 & 187.0 & 3.9 & 32.5 \\
Full \arscolor{ARS}: primary agent & 19.2 & 793.6 & 17.1 & 41.4 \\
Full \arscolor{ARS}: total & 24.9 & 980.6 & 21.1 & 39.3 \\
\midrule
\arscolor{ARS} w/o context separation & 19.0 & 1,097.2 & 15.2 & 57.6 \\
\bottomrule
\end{tabular*}
\end{center}
\end{table}

We also evaluate all three ablated workflows on the semantic-mismatch benchmark. On these short trajectories (31.7 frames on average), all three variants achieve scores close to those of full ARS across all three metrics (Table~\ref{tab:ablations-mismatch-workflow}). This simple benchmark therefore provides limited separation among the evaluated workflows. On LIBERO-Long, all three ablations yield lower downstream policy success than full ARS, supporting the utility of the complete workflow in this setting.

\begin{table}[ht]
\caption{Additional workflow ablations on the semantic-mismatch benchmark using Qwen3.6-27B.}
\label{tab:ablations-mismatch-workflow}
\begin{center}
\normalsize
\setlength{\tabcolsep}{4pt}
\begin{tabular*}{\linewidth}{@{\extracolsep{\fill}}lccc@{}}
\toprule
Configuration & VOC $\uparrow$ & FPSR $\uparrow$ & CMPG $\uparrow$ \\
\midrule
\arscolor{ARS} w/o verification \& inspection & 0.991 & 0.915 & 0.895 \\
\arscolor{ARS} w/o verification & 0.995 & 0.916 & 0.910 \\
\arscolor{ARS} w/o context separation & 0.993 & 0.913 & 0.910 \\
Full \arscolor{ARS} & 0.995 & 0.918 & 0.904 \\
\bottomrule
\end{tabular*}
\end{center}
\end{table}

\subsubsection{GVL-style single-prompt evaluation}
\label{app:gvl-single-prompt}

\paragraph{Purpose and protocol.}
We evaluate Qwen3.6-27B with a GVL-style single prompt \citep{gvl} to assess the progress-estimation capabilities of the base model under direct prompting. The no-inspection ARS variant still permits multi-turn reasoning and harness checks, whereas GVL-style inference produces its progress estimates in a single model response. We provide the task instruction and all recorded trajectory frames, without temporal subsampling. The initial frame serves as the reference scene, and the model predicts task-completion percentages for the remaining frames. We evaluate both chronological presentation and the shuffled presentation used by GVL. In the shuffled variant, the initial reference frame remains fixed and the remaining frames are randomly permuted. Predictions are mapped back to their original temporal order before evaluation and action-chunk selection.

\paragraph{Semantic-mismatch benchmark.}
Chronological presentation performs better than shuffled presentation in our experiments. We therefore report the chronological variant in Table~\ref{tab:ablations-available}, which achieves 0.637 VOC, 0.900 FPSR, and 0.354 CMPG. Although its FPSR approaches that of ARS (0.918), its lower VOC and CMPG indicate weaker progress ordering and separation between correct and mismatched instructions. Thus, suppressing progress increases under mismatched instructions alone does not establish the overall quality of the estimated progress signal.

\paragraph{LIBERO-Long.}
We apply the same action-chunk selection rule as in the main simulation experiments, retaining chunks with $\Delta_t = p_{t+H} - p_t > 0$, where $H=10$. Table~\ref{tab:gvl-ordering-libero} reports policy success and the percentage of action chunks retained for training. With chronological presentation, 95.20\% of chunks have zero progress change and only 3.58\% are retained. Policy training completes, but the resulting policy achieves 0.0\% success.

Shuffling reduces the zero-change fraction to 70.50\% and increases the retained fraction to 14.80\%. The resulting policy achieves 34.1\% success, 10.1 percentage points below Vanilla BC and 23.8 points below ARS with the same Qwen3.6-27B base model. These results show that frame ordering substantially affects the progress signal produced by single-prompt inference. However, the additional variation induced by shuffling does not translate into effective data selection relative to unfiltered behavior cloning in this setting.

\begin{table}[ht]
\caption{GVL-style single-prompt inference with Qwen3.6-27B on LIBERO-Long, using all trajectory frames. Retained is the percentage of action chunks kept for policy training.}
\label{tab:gvl-ordering-libero}
\begin{center}
\begin{tabular}{lcc}
\toprule
Frame order & Success (\%) $\uparrow$ & Retained (\%) \\
\midrule
Chronological & 0.0 & 3.58 \\
Shuffled & $34.1 \pm 1.1$ & 14.80 \\
\bottomrule
\end{tabular}
\end{center}
\end{table}

\subsection{The form of intermediate reasoning}
\label{app:intermediate-reasoning}

R\textsuperscript{2}VLM and SOLE-R1 organize intermediate reasoning primarily through explicit language CoT. R\textsuperscript{2}VLM recurrently maintains task decomposition, key steps, and their completion status through CoT over ordered video snippets \citep{r2vlm}. SOLE-R1 generates spatiotemporal CoT before each progress prediction; removing CoT from its progress training produces flatter, noisier rewards and weakens downstream learning \citep{soler1}.

\begin{table}[ht]
\caption{Effect of the base model's thinking mode under the same ARS workflow with Qwen3.6-27B. LIBERO-Long uses no outcome conditioning; Retained is the percentage of chunks kept from the original dataset.}
\label{tab:ablations-thinking}
\begin{center}
\normalsize
\setlength{\tabcolsep}{4pt}
\begin{tabular*}{\linewidth}{@{\extracolsep{\fill}}lccccc@{}}
\toprule
\multirow{2}{*}{Thinking mode} & \multicolumn{3}{c}{Semantic-mismatch} & \multicolumn{2}{c}{LIBERO-Long} \\
\cmidrule(lr){2-4}\cmidrule(l){5-6}
 & VOC $\uparrow$ & FPSR $\uparrow$ & CMPG $\uparrow$ & {Success (\%) $\uparrow$} & {Retained (\%)} \\
\midrule
Disabled & 0.990 & 0.915 & 0.877 & 55.4 $\pm$ 0.7 & 62.0 \\
Enabled & \textbf{0.995} & \textbf{0.918} & \textbf{0.904} & \textbf{57.9} $\pm$ 0.4 & 66.9 \\
\bottomrule
\end{tabular*}
\end{center}
\end{table}

We share their emphasis on intermediate reasoning, but examine whether it must take this form. Disabling the base model's thinking mode largely preserves ARS's semantic-mismatch performance and yields 55.4\% downstream success, 11.2 points above Vanilla BC (44.2\%). Enabling thinking further improves success by 2.5 points to 57.9\% (Table~\ref{tab:ablations-thinking}). The harness still maintains and revises reasoning state through its event timeline, evidence links, and progress curve. These results suggest that a standalone language CoT trace is not the only effective way to organize intermediate reasoning for reward estimation.

\section{Additional Details on Real-World Policy Learning}
\label{app:realrobot-details}

\subsection{Task and evaluation protocol}

The robot must complete three screw-fastening operations on the washing machine's rear panel. Success is evaluated at the complete-task level: all three screws must be successfully fastened, as confirmed by a human evaluator. The system monitors end-effector rotation time and torque, and automatically marks timed-out trials as failures. Avoiding a timeout does not by itself establish success; final completion is determined by human assessment.

Real-robot evaluation is substantially more time-consuming than simulation. We therefore use one training seed per configuration and evaluate each policy over 200 complete-task trials. These results do not quantify variability across training seeds.

\subsection{Policy training and data sampling}

We initialize all policies in Table~\ref{tab:realrobot} from the same pretrained $\pi_{0.5}$ checkpoint \citep{pi05}. We use an action-chunk horizon of $H = 50$ and evaluate the checkpoint saved at 200,000 gradient steps for all configurations.

The dataset contains 2,574 demonstration trajectories and 3,092 offline rollout trajectories. Demonstration-only BC uses all demonstration chunks. The remaining configurations combine the demonstrations with either all rollout chunks or the subset selected using ARS progress estimates. ARS-based configurations use the stage-dependent selection rule specified in Section~\ref{sec:progress-guided-policy-learning}.

For every configuration that incorporates rollout data, we use stratified sampling with a fixed 1:1 ratio between demonstration and rollout chunks. Within each source, eligible chunks are sampled uniformly. For ARS-based configurations, the rollout pool consists of the retained chunks. Thus, filtering changes the size and composition of the rollout pool while preserving the relative sampling weights of demonstrations and rollouts. We apply the BC loss in Equation~\ref{eq:bc} separately to each source, with equal expected weights under this sampling scheme.

\paragraph{Visual references.}
We provide two positive examples showing initial end-effector engagement with the screw head and full tightening with end-effector withdrawal, and one negative example showing stalled fastening. Each reference consists of a single image and a brief description, without numerical progress annotations. The references are supplied during ARS progress inference.

\subsection{Outcome conditioning on the real robot}
\label{app:realrobot-outcome}

Adding terminal outcome labels to ARS with visual references increases observed success from 81.5\% to 84.0\%, but the incremental gain is not statistically significant (two-sided Fisher's exact test, $p=0.597$).

\end{document}